\documentclass[acmsmall]{acmart}
\AtBeginDocument{%
  \providecommand\BibTeX{{%
    \normalfont B\kern-0.5em{\scshape i\kern-0.25em b}\kern-0.8em\TeX}}}

\setcopyright{acmcopyright}
\acmDOI{XXXXXXX.XXXXXXX}

\usepackage{lineno}
\usepackage{natbib}
\usepackage{amsmath,bm,float,graphicx,multirow,soul,xspace}
\usepackage{bbm} 
\usepackage{booktabs} 
\usepackage{hyperref} 
\usepackage[mathscr]{eucal}
\usepackage{subcaption} 
\usepackage{wrapfig} 

\begin{document}




\title{Identifying Health Risks from Family History: A Survey of Natural Language Processing Techniques}




\author{Xiang Dai}
\affiliation{%
  \institution{CSIRO Data61}
  \city{Sydney}
  \country{Australia}}
\email{dai.dai@csiro.au}

\author{Sarvnaz Karimi}
\affiliation{%
  \institution{CSIRO Data61}
  \city{Sydney}
  \country{Australia}}
\email{sarvnaz.karimi@csiro.au}

\author{Nathan O’Callaghan}
\affiliation{%
  \institution{CSIRO Health \& Biosecurity}
  \city{Adelaide}
  \country{Australia}}
\email{nathan.o'callaghan@csiro.au}

\renewcommand{\shortauthors}{Dai, Karimi, and O’Callaghan}

\begin{abstract}
Electronic health records include information on patients' status and medical history, which could cover the history of diseases and disorders that could be hereditary. One important use of family history information is in {\em precision health}, where the goal is to keep the population healthy with preventative measures. Natural Language Processing (NLP) and machine learning techniques can assist with identifying information that could assist health professionals in identifying health risks before a condition is developed in their later years, saving lives and reducing healthcare costs.

We survey the literature on the techniques from the NLP field that have been developed to utilise digital health records to identify risks of familial diseases. We highlight that rule-based methods are heavily investigated and are still actively used for family history extraction. Still, more recent efforts have been put into building neural models based on large-scale pre-trained language models.
In addition to the areas where NLP has successfully been utilised, we also identify the areas where more research is needed to unlock the value of patients' records regarding data collection, task formulation and downstream applications.
\end{abstract}









\maketitle



\section{Introduction}





Improving the health of the population over the years and reducing their likelihood of developing diseases could help reduce healthcare expenditures, improve life expectancy, and guarantee a healthier population when they reach their older years~\cite{harris-2018-aged-care}. Family history is one factor that, if investigated and taken into account for clinical decision-making, could prevent further complications in people's lives. That is, by putting into place plans where potential risk is reduced for developing a serious condition, such as colon cancer, not only is health expenditure reduced, but the quality of life is improved for those at risk~\cite{brenner-stock-2014-colorectal-cancer,kaphingst-utah-2021-bmc-genetics}. 

There are a number of diseases and disorders---such as heart diseases or dementia---that are considered familial or hereditary, that is, having those in the biological family could increase the risk of developing it~\cite{barrett-connor-1984-family-history,murff-byrne-2004-family-history}. Identifying these risks from family history stored in Electronic Health Records (EHR)---either in a structured format or as free-text notes---is an area of interest. In this survey, we review how Natural Language Processing (NLP) techniques help with identifying health risks from documented family history. Published literature on family history extraction includes identifying whether parts of a patient's clinical note are about their family history~\cite{bill-minnesota-2014-amia-family-history}; extracting family members and observations from clinical narratives~\cite{zhou-lu-2014-family-relatives,shen-mayo-2021-jmir-n2c2-2019-fh}; and mining attributes such as age of onset, uncertainty, side of family~\cite{rama-oslo-2018-louhi-family-history}.

\paragraph{What is family history, and why is it useful?}
The value of family history information has long been recognised~\cite{fuchs-giovannucci-1994-family-history,scheuner-wang-1997-family-history,guttmacher-collins-2004-family-history,del-fiol-utah-2020-clinical-decision-support}.
Family history can be used to estimate the risk of a patient developing certain diseases such as breast cancer, colorectal cancer, ovarian cancer, osteoporosis, cardiovascular disease, psychiatric disorders, and diabetes~\cite{polubriaginof-columbia-2015-amia-family-history,wang-mayo-2017-amia-family-history}.
It can also signal the need for further genetic counselling or preventative health screening~\cite{bill-minnesota-2014-amia-family-history}.
For example, Permuth-Wey et al.~\cite{permuth-egan-2009-familial-cancer} find that individuals with a family history of pancreatic cancer have nearly a two-fold increased risk for developing pancreatic cancer to those without such a history.
They suggest that families with two or more pancreatic cancer cases may benefit from comprehensive risk assessment using detailed family history data and that those at highest risk may be referred to screening programs.
Different from a genetic susceptibility that can be obtained via genetic profiling~\cite{patel-gonen-2012-genetic-profiling}, family history information captures the complex interactions between genetic, environmental, and behavioural factors~\cite{roberts-fushman-2020-trec-precision-medicine}.
Milne et al.~\cite{milne-auckland-2009-family-hisory} study the associations between family history of four psychiatric disorders---depression, anxiety, alcohol dependence and drug dependence---and clinical features of these disorders in probands (the first diagnosed family member). They suggest that family history is useful for determining patients' clinical prognosis and for selecting cases for genetic studies.
Additionally, Claassen et al.~\cite{claassen-vumc-2010-bmc-family-history} point out that family history information can be used to personalize health messages, which motivate at-risk individuals to adopt and maintain healthy lifestyles in order to prevent disease.

\subparagraph{A motivating example}
Pancreatic cancer is the fourth leading cause of cancer death in the United States, while an estimated 466,003 people died from it worldwide in 2020.\footnote{\url{https://www.cancer.net/cancer-types/pancreatic-cancer/statistics}}
Due to its few signs and symptoms until in well-advanced cancer stages, pancreatic cancer is referred to as a silent killer: around 73\% of patients die within the first year of their diagnosis~\cite{mehrabi-mayo-2015-family-history}.
Several risk factors, such as family history, pancreatic cysts, obesity, smoking, and alcohol intake, have been found.
It is estimated that 10\% of pancreatic cancers have a familial basis.
Risk is increased 7-9 fold if one first-degree relative (parents, siblings or children) has pancreatic cancer.

\paragraph{Why we need NLP for extracting family history}

Modern healthcare systems have enabled structured information about family history to be filled out through paper forms or online programs. However, Taber et al.~\cite{taber-utah-2020-jamia-family-history} find that the dedicated electronic module for family history is poorly suited to clinical workflows, and a considerable fraction of family history information is contained in free-text clinical notes, which are written by clinicians or genetic counsellors after they encounter with patients (An example can be found in~\ref{quote-example-family-history}).
For example, the Epic EHR includes a dedicated portion for collecting information on family history.
The portion allows for structured entry of problems (selected from a list of 210 values such as cancer and diabetes), family members (selected from a list of 21 values such as mother and brother), age of onset (a numeric value such as 68), and a free-text field called \emph{comment}.
Chen et al.~\cite{chen-vermont-2012-amia-family-history} conduct a manual review on $3,358$ unique comments to identify reasons for use of the comment field and what types of information are contained within them.
They find that greater than one-third of these comments are used to collect information that should be entered into available structured fields (a problem that can be selected from the list of 210 values or a specific onset age, such as `breast cancer at 70').
Polubriaginof et al.~\cite{polubriaginof-columbia-2015-amia-family-history} also assessed the quality of family history data captured in an established commercial EHR at a medical centre. 
After analysing differences between 10,000 free-text and 9,121 structured family history observations, they found that free-text notes contain more information than structured notes.

\begin{quote}

The patient has a 1 3/4-year-old daughter who is in good health.  She has had three prior miscarriages, the last of which had normal chromosomes.  The patient has one 23-year-old brother who is in good health.  He has a 1 3/4-year-old daughter who is also healthy.  The patient's mother died at the age of 34 due to syndrome hemiparaplegic.  She was diagnosed in her late 20’s.  The patient's father died at 44 of an imbecile.  He had enamel caries earlier in his life.

The patient's husband is 56 and in good health.  He has two sisters, ages 31 and 54.  His oldest sister has six healthy children.  His younger sister has no children.  His mother is 45 and has down syndrome.  His father is 45 and healthy.

\label{quote-example-family-history}
\end{quote}

In this survey, we study how NLP can be used to extract family history information. After We review NLP tasks, data resources, and methods of family history extraction, we provide a discussion about several research directions.

\paragraph{Study selection}
We start from the article summarizing the 2019 National NLP Clinical Challenges (n2c2)/Open Health Natural Language Processing (OHNLP) Competition on family history extraction~\cite{shen-mayo-2021-jmir-n2c2-2019-fh} and find relevant articles via the citation graph.
That is, once we identify one \emph{relevant} study---describing either NLP techniques of extracting family history from free text or family history datasets that enable the development of NLP models, we retrieve its citation and references using Semantic Scholar Academic Graph API,\footnote{\url{https://www.semanticscholar.org/product/api}} and then the relevance for each retrieved article is manually decided. This process is repeated until no new relevant article is identified. In total, we retrieve $577$ articles, $25$ out of which are relevant.

\section{Tasks and Resources}

\begin{figure}[t]
    \centering
    \includegraphics[width=0.6\linewidth]{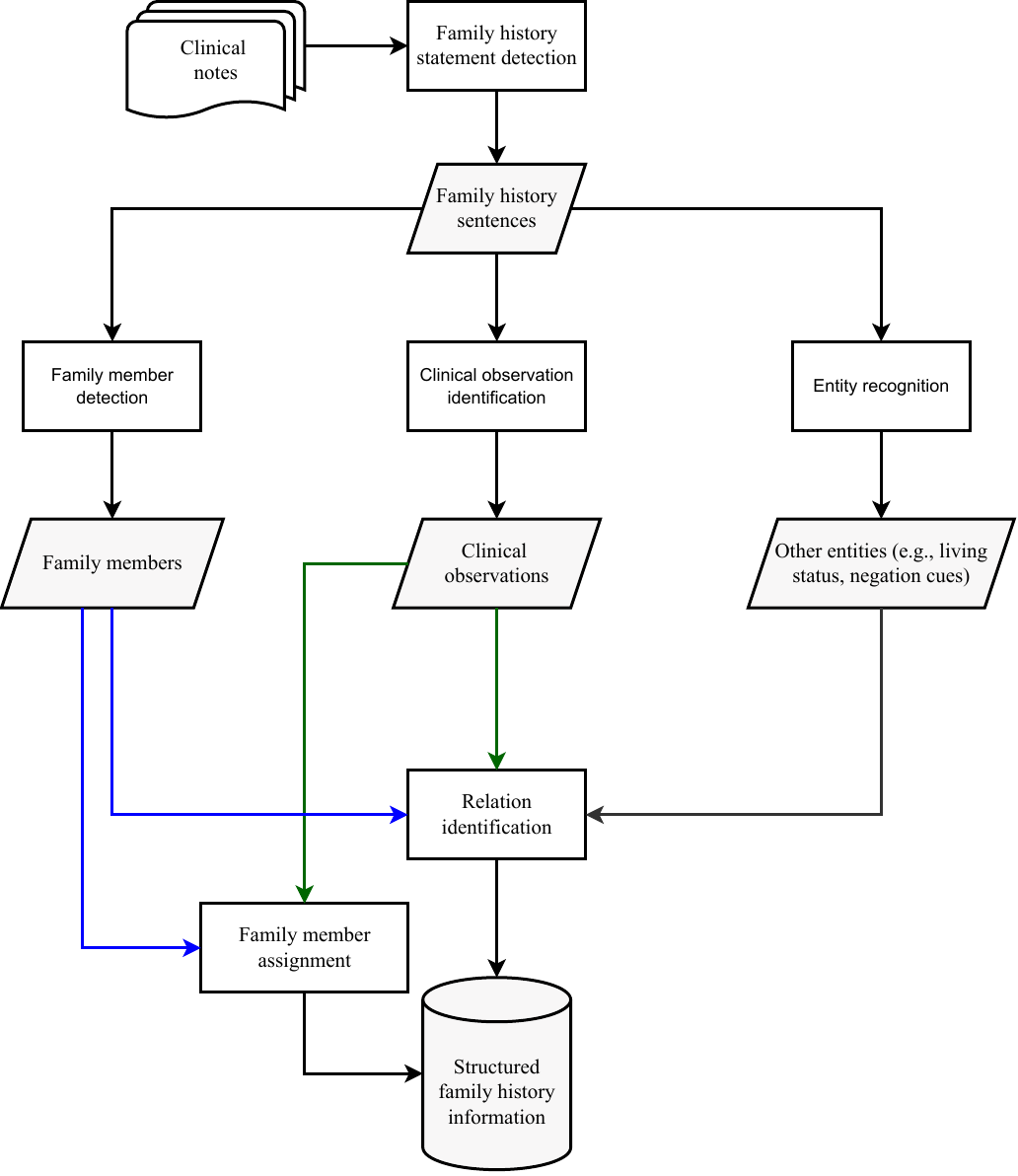}
    \caption{The main tasks, represented in rectangles, in the family history extraction pipeline.\label{figure_tasks_studied}}
\end{figure}

In this section, we summarise what NLP tasks have been addressed in terms of analysing family history and what resources have been available to enable the development of NLP models for family history extraction. A pipeline of main tasks is shown in Figure~\ref{figure_tasks_studied}.

\subsection{Tasks}
\label{section-task}

\paragraph{Family history statement detection}


While it is common to have a dedicated section of family history in commercial EHR, Lewis et al.~\cite{lewis-ibm-2011-ieee-parsing-family-history} found that only 36\% of family history information located within these dedicated sections. In other words, family history may be buried in various places in the clinical text and usually mixed with the patient's medical history.
Therefore, the first component in a family history extraction pipeline is usually to determine if one sentence contains a family history statement~\cite{bill-minnesota-2014-amia-family-history}.

\paragraph{Family member detection}
The simplest format of family member detection task is to identify specific words, e.g., `brother', `son', in text that refer to the relative of the patient. %
However, there are several constraints that make the task less straightforward. %
The first one is that non-blood relatives, e.g., `adopted siblings', `partner's sister', are usually excluded in previous work as their information are not helpful to analyse heritage diseases that the patient may have~\cite{shen-mayo-2021-jmir-n2c2-2019-fh,balabaeva-kovalchuk-2020-family-disease-tree}. %
The second constraint is that the possible surface forms of mentioned family member need to be normalized with the side of family. For example, the family member mentioned in `The mother has a brother who is an alcoholic and ...' is not the `brother' of the patient but the `uncle', and the side of family is `maternal'. %
Lastly, information about two or more family members may be mentioned together. For example, there are two family members mentioned in the sentence `both cousins suffered from high blood pressure', and there is a need to distinguish them because they may not have exactly the same health conditions. %
In addition to family member mention, previous work also attempts to detect their attributes, such as living status that show health status of the family members~\cite{shen-mayo-2021-jmir-n2c2-2019-fh}.

\paragraph{Clinical observation identification}
This task involves identifying any health-related problems, which are defined in varied ways in previous studies.
For example, Shen et al.~\cite{shen-mayo-2021-jmir-n2c2-2019-fh} aim to identify problems including diseases, smoking, suicide and drinking, excluding auto accident, surgery, and medications.
Rama et al.~\cite{rama-oslo-2018-louhi-family-history} distinguish between clinical event (something happens and then is over, for example, the patient has a heart attack) and clinical condition (a prolonged state, for example, the patient has heart disease).
They describe a range of conditions such as diseases, diagnoses, various types of mutations, test results, treatments, and vital state.
Lewis et al.~\cite{lewis-ibm-2011-ieee-parsing-family-history,lewis-sfsu-2011-bicob-family-history} consider diseases that must be resolvable to a 3 digit ICD9 (International Classification of Diseases, Ninth Revision) code.
In addition to recognize clinical observations (also known as \emph{concept identification}) in text, previous work also attempt to map from the surface form of these observations to their standardised names in the SNOMED CT, which is also known as \emph{concept normalization}~\cite{dai-csiro-2021-cikm-searchehr}, and to detect negation property for these observations~\cite{shen-mayo-2021-jmir-n2c2-2019-fh,rama-oslo-2018-louhi-family-history}.

\paragraph{Relation identification}
The most common type of relation considered by previous work is the relation between clinical observation and its family member holder, and the task is also known as family member assignment~\cite{shen-mayo-2021-jmir-n2c2-2019-fh,dai-csiro-2021-cikm-searchehr}.
Rama et al.~\cite{rama-oslo-2018-louhi-family-history} define a number of other relations between entities. For example, there is `Related to' relation between family members, and a more specific relation `Partner' is used to identify couples that are able to provide offspring.
Similarly, Yang et al.~\cite{yang-florida-2020-jmir-family-history} split the relation identification task as a two-step task: 1) determine whether there is a relation between any two entities (i.e., in-relation or no-relation); and 2) classify the correct relation type based on the entity type (e.g., family member---observation relation and family member---living status relation).

\paragraph{Other tasks using NLP}

In addition to family history extraction, there are other works that use NLP techniques to study the characteristics of family history recorded in EHRs~\cite{chen-vermont-2012-amia-family-history,mowery-kawamoto-2019-amia-family-history} or integrate family history into a downstream application\cite{wan-ge-2022-family-history-mood-disorder,shi-utah-2022-jmir-family-history}. 

Wan et al.~\cite{wan-ge-2022-family-history-mood-disorder} classify family history in admission notes from a Chinese psychiatric medical center into five subgroups: schizophrenia, depressive episodes, bipolar disorder, other non-organic mental disorders, and unspecified non-organic psychosis.
The main goal of their work is to study onset characteristics among patients with a mood disorder.
Chen et al.~\cite{chen-vermont-2012-amia-family-history} focus on free-text comments within the structured family history section and aim to identify reasons for the use of the comments field. 
Mowery et al.~\cite{mowery-kawamoto-2019-amia-family-history} explore ways of augmenting structured family history data using free-text comments when some information (e.g., age of onset and age of death) are not available as structure data.
Shi et al.~\cite{shi-utah-2022-jmir-family-history} augment clinical decision support algorithm~\cite{del-fiol-utah-2020-clinical-decision-support} that relies exclusively on structure family history data with free-text comments, with a focus on identifying patients who meet the criteria for genetic testing for hereditary cancers.

\subsection{Resources}
Most of the published studies on family history use in-house data, which are not openly available. For example, Friedlin et al.~\cite{friedlin-mcdonald-2006-amia-family-history} use admission notes dictated at a primary care hospital in Indianapolis, Indiana. Goryachev et al.~\cite{goryachev-harvard-2008-amia-family-history} use discharge summaries and outpatient notes from two hospitals in Boston, Massachusetts. 
Chen et al.~\cite{chen-vermont-2012-amia-family-history} use EHRs from a tertiary care academic medical center affiliated with the University of Vermont. 
Both Mowery et al.~\cite{mowery-kawamoto-2019-amia-family-history} and Shi et al.~\cite{shi-utah-2022-jmir-family-history} use EHRs for patients who visited the University of Utah Health primary care clinic. %
More recently, openly available resources start to attract attention as they allow interoperability and reproducibility.
We describe these openly available resources and provide a summary of these resources in Table~\ref{table-data-resource-summary}. %
A further discussion about data collection can be found in Section~\ref{section-future-data-collection}.

MTSample dataset, used in~\cite{bill-minnesota-2014-amia-family-history,lewis-ibm-2011-ieee-parsing-family-history,lewis-sfsu-2011-bicob-family-history}, is an openly available resource that contains sample transcription reports for 40 specialties (e.g., general medicine, cardiovascular, urology). These reports have been anonymised and submitted by transcriptions and clinicians, and they can be used as a proxy of the kinds of texts generated in a clinical setting.
It is worthy noting that MTSample is not a dedicated resource for family history, the family history statements are thus relatively scarce.
For example, Lewis et al.~\cite{lewis-ibm-2011-ieee-parsing-family-history} find that, out of 3940 clinical reports, there are only 1220 unique sentences mentioning at least one family member.
Similarly, Bill et al.~\cite{bill-minnesota-2014-amia-family-history} find that only $284$ out of $23,155$ sentences contain family history statements.

To standardize evaluation on family history extraction, Mayo Clinic researchers organize the BioCreative/OHNLP family history extraction shared task in 2018~\cite{rastegar-mayo-2018-bcb-biocreative-2018} and then the n2c2/OHNLP shared task in 2019~\cite{shen-mayo-2021-jmir-n2c2-2019-fh}. %
The shared tasks focus on identifying family members and clinical observations, and extracting the association of the living status, side of the family and clinical observations with family members. %
They curate the deidentified clinical narrative from family history sections of clinical notes at Mayo Clinic Rochester. %
That is, they manually removed all the patient-protected information (e.g., names, locations, and age above 89) and shuffle the observations, family members, and ethnicities among the whole corpus to further protect the confidentiality.
Note that automatically generated semi-structured texts and sections that combine the patients' social history with the family history are excluded.
The corpus used in~\cite{shen-mayo-2021-jmir-n2c2-2019-fh} is annotated by a 5-member annotator team, and each document is annotated by 2 annotators. %
One coordinator worked as the adjudicator to resolve discrepancies. %

Rama et al.~\cite{rama-oslo-2018-louhi-family-history} develop the annotation guidelines for family history extraction using synthetically produced clinical text in an iterative fashion. That is, the domain expert participates both in the annotation and development of guidelines. Synthetic examples are generated to challenge the annotation guidelines and then the guidelines are revised based on the annotation disagreements. %
In addition to `Family' members, Rama et al.~\cite{rama-oslo-2018-louhi-family-history} use `Self' to refer the patient under consideration, and `Index' to the proband---the first identified family member with a genetic disorder. Medical concepts---diseases, diagnoses, mutations, test results, treatments and vital state---are grouped into one entity category called `Condition'; `Event' (e.g., has a heart attack) is distinguished from `Condition' (e.g., have heart disease); a set of modifiers---side of the family, age, negation cues, amount, temporal---are annotated to assist semantic interpretation of family history information. Rama et al.~\cite{rama-oslo-2018-louhi-family-history} also define several relations between entities: holder (between Condition and Family/Self/Index entities); modifier (between modifiers and other entities); related to (between family members); subset (used when multiple family members are mentioned at the same time, e.g., between `two brothers' and `one of them'); partner (between no kinship family members that are able to provide offspring, e.g., between `husband' and `wife'). %

Azab et al.~\cite{azab-umich-2019-emnlp-family-history} collect natural language interactions---answers to a family history questionnaire---and then annotate medical family histories. The questionnaire consists of questions targeting the medical history of the patient and their relatives up to third degree, and answers are collected from volunteers through crowd-sourcing or test cases of genetic counseling sessions. Finally, a genetic counseling expert manually annotate family members, their attributes (e.g., name, gender, age), and their illnesses. %

\begin{table}[t]
    \centering
    \footnotesize
    \begin{tabular}{p{0.21\linewidth} p{0.21\linewidth} p{0.5\linewidth}}
    \toprule
    \bf Data & \bf Size & \bf Description \\
    \midrule
    n2c2/OHNLP 2019~\cite{shen-mayo-2021-jmir-n2c2-2019-fh} & 216 reports & Deidentified clinical narrative from clinical notes \\
    \hline
    UMich~\cite{azab-umich-2019-emnlp-family-history} & 4,304 sentences & Answers to a family history questionnaire; focus on Cancer \\
    \hline
    UiO~\cite{rama-oslo-2018-louhi-family-history} & 477 sentences & Synthetically produced clinical text; focus on Cardiac disease \\
    \hline
    MTSamples~\cite{bill-minnesota-2014-amia-family-history,lewis-ibm-2011-ieee-parsing-family-history,lewis-sfsu-2011-bicob-family-history} & 491 $\sim$ 3940 reports & Sample medical transcription reports \\
    \bottomrule
    \end{tabular}
    \caption{A summary of openly available resources.}
    \label{table-data-resource-summary}
\end{table}

\section{Methods}

Techniques for family history extraction have developed gradually from rule-based to machine learning-based and deep learning-based during the last several decades.
In this section, we describe representative work and focus for identifying development trends.

\subsection{Rule-based approach}
Early studies of family history extraction are dominated by rule-based methods, where rules are developed either based on external resources such as biomedical ontologies or vocabularies, or by clinical experts and linguists based on the analysis of language patterns in clinical text~\cite{mehrabi-mayo-2015-family-history}.

Mehrabi et al.~\cite{mehrabi-mayo-2015-family-history} develop a rule-based algorithm based on the SecTag terminology---a large scale effort to assemble a list of terminologies used as section headers---to identify the clinical notes sections for family history.
In contrast, Lewis et al.~ \cite{lewis-sfsu-2011-bicob-family-history} identifies candidate family history sentences based on the presence of `sign-post' words (e.g., `mother', `brother'), because they aim to extract family histories from any section (rather than only titled sections of family history) within a clinical report.

Friedlin et al.~\cite{friedlin-mcdonald-2006-amia-family-history} develop a rule-based system for extracting family history for $12$ diseases and categorizing relative degrees (primary or secondary).
They first map noun phrases to the UMLS concepts of family member and diagnosis~\cite{bodenreider-2004-umls}, and then use a set of co-occurrence based rules to associate diagnoses to family members.
Bill et al.~\cite{bill-minnesota-2014-amia-family-history} rely on two external resources for family member detection and observation recognition.
They search for matches to phrases in the HL7 Clinical Genomics family history model for family member detection and phrases defined by the UMLS Semantic Groups as disorders and procedures for observation recognition.
The identified family member and observation entities are then linked together by an indicator word or phrase thus forming a predication relation.
A lexicon of indicator words and phrases, including possession words (e.g., has, is, with) and experience indicators (e.g., suffered a, died of, recovered from), is constructed from the annotated training set.

Almeida et al.~\cite{almeida-matos-2020-sac-family-history} manually compile a lexicon including all family members and use the co-reference graph to add the corresponding annotations to pronouns. 
They also compile a small lexicon extracted from the training data of n2c2/OHNLP shared task~\cite{shen-mayo-2021-jmir-n2c2-2019-fh} to identify living status, and a dictionary from the UMLS Metathesaurus~\cite{bodenreider-2004-umls} for clinical observation identification.
Finally, they follow the shortest path in the dependency graph, which is built using Stanford CoreNLP~\cite{manning-stanford-2014-acl-corenlp}, to associate clinical observations to family members.
Lewis et al.~\cite{lewis-sfsu-2011-bicob-family-history} also employs a set of dependency-based syntactic patterns to link a family member to its diagnosis. For example, if a sentence contains a direct object dependency (\emph{dobj}) and a nominal subject dependency (\emph{nsubj}), and both dependencies share the same governor, their method recognizes the dependent of the \emph{dobj} dependency as disease and the dependent of the \emph{nsubj} dependency as family member. %
Similarly, Lewis et al.~\cite{lewis-ibm-2011-ieee-parsing-family-history} compile 107 unique dependency paths (e.g., `mother' $\to$ \emph{nsubj} $\to$ `has' $\to$ \emph{dobj} $\to$ `cancer') from 225 family history sentences.
They find that the most frequent 10 dependency paths can recover 57\% of the family history tuples.

It is worthy noting that rule-based methods are still actively used for family history extraction in the literature, but many recent work use off-the-shelf NLP tools which are actually machine learning-based (e.g., Stanford CoreNLP~\cite{manning-stanford-2014-acl-corenlp}) or even deep learning-based (e.g., Stanza~\cite{zhang-stanford-2021-jamia-stanza}).
One potential limitation of these work is, most of the time, off-the-shelf tools are taken for granted and their effectiveness on clinical text is not well studied. 
In other words, most of these work assume these tools are effective at processing clinical text (e.g., for syntactic parsing) and aim to design task specific rules to extract family history based on outputs from these NLP tools.

\subsection{Statistical machine learning approach}
Instead of applying `hard' rules to extract family history, researchers also investigate machine learning-based method that uses ‘soft’ features and estimates the importance (weights) of features using labelled training data. Bill et al.~\cite{bill-minnesota-2014-amia-family-history} use a classification approach for family history statement detection.
They first use regular expressions to identify the section and subsection boundaries and headings.
Then, the model takes the section heading, subsection heading and sentence text as input together, and N-gram-tokenized text is used as predictors to a support vector machine classifier.

Rama et al.~\cite{rama-oslo-2018-louhi-family-history} and Azab et al.~\cite{azab-umich-2019-emnlp-family-history} model entity (family member and observation) recognition as a sequence labeling problem.
The model takes features corresponding to individual word---lexical features of current and surrounding words, POS tags of current and surrounding words, and entity types of previous words---as input, and predict entity type for each word using either a linear SVM model~\cite{rama-oslo-2018-louhi-family-history} or a CRF model~\cite{azab-umich-2019-emnlp-family-history}.
Additionally, Rama et al.~\cite{rama-oslo-2018-louhi-family-history} uses a SVM model---based on lexical features, POS tags, dependency features, and entity types---for relation classification. 

\subsection{Deep learning-based approach}
Despite the promising results of machine learning-based approach, its main shortcoming is the requirement of sophisticated feature templates, which is usually tailored to specific dataset.
To alleviate the burden of manually building feature templates, deep learning models enable automated feature extraction. 
\cite{dai-csiro-2021-cikm-searchehr,yang-florida-2020-jmir-family-history,dai-2019-bmc-family-member,dai-nkust-2020-jmir-family-history,rybinski-csiro-2021-jmir-family-history,kim-musc-2021-jmir-family-history,shi-hit-2019-bmc-family-history} formulate family member detection and clinical observation identification as a sequence tagging problem, and neural sequence models are employed to handle the problem.
To overcome the limitation of the small size of the training data, Dai et al.~\cite{dai-csiro-2021-cikm-searchehr} and Rybinski et al.~\cite{rybinski-csiro-2021-jmir-family-history} exploit domain-adaptive pre-training~\cite{gururangan-allenai-2020-acl-dapt} and intermediate-task pre-training~\cite{pruksachatkun-nyu-2020-acl-intermediate-task}, and they demonstrate a cumulative value of both techniques on clinical observation identification using BERT-CRF model~\cite{devlin-google-2019-naacl-bert}.
On family member detection, Rybinski et al.~\cite{rybinski-csiro-2021-jmir-family-history} find that BERT-CRF model achieves competitive results against a hybrid method that enhances a rule-based system with BERT-based paragraph filter.
Rybinski et al.~\cite{rybinski-csiro-2021-jmir-family-history} also experiment with a neural coreference resolution system to exploit cross-sentence context, but find that it does not improve the model effectiveness.

It is worthy noting that most of these studies still apply rule-based methods to determine the family role and to determine the family side~\cite{dai-csiro-2021-cikm-searchehr,rybinski-csiro-2021-jmir-family-history,silva-rafael-2020-jmir-family-history}, except in~\cite{yang-florida-2020-jmir-family-history,dai-nkust-2020-jmir-family-history}.
Yang et al.~\cite{yang-florida-2020-jmir-family-history} adopt a pure deep learning-based solution where they build different BERT-based classifiers to determine the family roles, family sides, negation of observations, and living status scores, respectively.
Dai et al.~\cite{dai-nkust-2020-jmir-family-history} propose an enhanced relation-side scheme that encodes the required family member properties. For example, they include the side and relationship information in the tag `E-Aunt-Paternal' to indicate the last token within a family member mention.
Yang et al.~\cite{yang-florida-2020-jmir-family-history,kim-musc-2021-jmir-family-history} create a voting ensemble method that combine the predictions of $10$ different Bi-LSTM models, that use dependency-based static word embeddings~\cite{komninos-manandhar-2016-naacl-dependency-embeddings} and ELMo embeddings~\cite{peters-allenai-2018-naacl-elmo} as input.

Zhan et al.~\cite{zhan-hit-2021-jmir-family-history} propose a graph-based model with biaffine attention to jointly extract entities and relations about family history.
They frame the family history extraction task as a dependency parsing problem. That is, tokens belonging to the same entity (e.g., family member, clinical observation, living status) are connected together by an `app' arc from right to left; entities with a relation (e.g., family member and clinical observation; family member and living status) are connected through linking the right most token by an arc labeled with the entity type; and, tokens that do not belong to any entities are connected with the `ROOT' node by `NULL' arcs.
Zhan et al.~\cite{zhan-hit-2021-jmir-family-history} first use BERT~\cite{devlin-google-2019-naacl-bert} and CNN-BiLSTM~\cite{ma-hovy-2016-acl-ner} to build contextual representations of each token in the sentence, and then use two biaffine classifiers~\cite{dozat-manning-2017-iclr-biaffine} for unlabeled arc prediction and arc label prediction, respectively.
Finally, a rule-based post-processing module is used to convert the output to entities and relations; normalize family members; determine the side of family members; and, detect negation information of observations.





\paragraph{Summary of technical trends}
Rule-based methods were heavily investigated and are still actively used for family history extraction.
The main reason is that the development of rule-based methods usually directly benefit from the incorporation of domain knowledge via existing resources and manually compiled rules.
This approach is effective especially when the task is well defined and the input text is well written in a consistent manner.
However, Goryachev et al.~\cite{goryachev-harvard-2008-amia-family-history} find that these rules fail on sentences with multiple observations or multiple family members, whereas human annotators agree perfectly on these annotations. 
On the other hand, deep learning-based methods, especially those based on large-scale pre-trained language models, start to attract interests due to its pre-training-then-fine-tuning paradigm.
This is a promising approach for family history extraction, because models can learn to understand text via pre-training on raw-text and then to solve the task via fine-tuning on task-specific annotated data.
However, one common type of errors is that the model tends to extract information about all family members that are not related by blood or too distant~\cite{rybinski-csiro-2021-jmir-family-history}, and the model may need more task-specific annotated data to learn the subtlety.

\section{Future Research Directions}
Based on the reviewing of published studies on family history extraction, we identify three future research directions regarding data collection, task formulation, and application deployment, respectively.

\subsection{How should we collect family history data?}
\label{section-future-data-collection}
The development of NLP models, especially models based on deep learning techniques, usually requires the access to a large amount of annotated data.
Preparing and releasing such data requires complex thoughts and actions~\cite{gao-wisc-2021-jamia-review-clinical-nlp}.
However, health-related data stored in EHRs are private and the access to these data is restricted, not to mention the data annotation bottleneck because of the domain knowledge required~\cite{spasic-nenadic-2020-jmir-clinical-review}.
On the other hand, Clift et al.~\cite{clift-mayo-2022-family-history} find that the more significant a patient's family history, the less likely it was to be recorded accurately and consistently in EHR.

Previous efforts of creating openly available datasets for family history extraction use either synthetic data~\cite{rama-oslo-2018-louhi-family-history,brekke-rama-2021-synthetic-data} or anonymized data~\cite{shen-mayo-2021-jmir-n2c2-2019-fh}.
There are several issues around this.
On one hand, it is unclear whether synthetic data, which are usually carefully curated, can be fully representative of real world clinical notes, which are written under time pressure.
There is very few studies investigating how do these NLP models trained on synthetic clinical text perform on real world clinical text. 
On the other hand, anonymized data, although very similar to real world clinical texts, is usually of very small size.
Huge amount of human efforts are required to make sure all personally identifiable information are removed from data~\cite{shen-mayo-2021-jmir-n2c2-2019-fh}.

Azab et al.~\cite{azab-umich-2019-emnlp-family-history} develop a dataset consisting of natural language interactions obtained during genetic consultation and via crowdsourcing.
Their research points out an interesting direction that a dialog agent may be able to help with collecting data from the patient, or even the crowd-worker.
Developing chat-bot that asks some generic questions and elicits narrative answers from the customer has attracted a lot of interests from both academic and industry.
Applications have emerged in different domains, including the healthcare domain~\cite{laranjo-mq-2018-jamia-chatbot}.
Deploying a dialog agent may have additional benefits in healthcare domain.
For example, deploying a dialog agent to collect family history enable both patients and doctors be away from pressured consultation environment.

We believe there are two important steps to make this direction more appropriate.
The first is dialog agent needs to learn to ask the proper question given the context (the reason of consultation, patient medical history and existing knowledge about particular diseases).
Analogically, if the patient goes to have the skin checked for melanoma, the doctor may ask questions about their family history of skin cancer (but not stomach cancer etc.), their job (indoor or outdoor environment), their exposure to sunshine, etc.
The second is we need to develop domain-agnostic techniques that make models to be robust on different types of text, e.g., the model trained on conversation data to be effective on clinical notes.

\subsection{How can we define an unified framework for family history extraction?}
As we have shown in Section~\ref{section-task}, published studies mainly solve family history extraction as either text classification task (e.g., identifying whether a sentence is describing family history) or relation extraction task (e.g., extracting tuples connecting family member and observations).
However, there is no consensus on how the data are annotated, and developed methods are thus tailored towards particular dataset used~\cite{rybinski-csiro-2021-jmir-family-history}.

To overcome the low-resource issue, Rybinski et al.~\cite{rybinski-csiro-2021-jmir-family-history} investigates transfer learning technique that uses publicly available resource that are related (but not identical) to family history extraction.
Their research raises us a fundamental question: how family history extraction task/dataset is different from other information extraction tasks/datasets? In other words, can we frame family history extraction in a generalisable manner so that the model can learn to solve the task by training on other related tasks.
Also, we believe a well-defined framework should be able to allow reusing of existing resources and benchmarking the progress in the area.

\begin{figure}[t]
\begin{subfigure}{\textwidth}
  \centering
  \includegraphics[width=\linewidth]{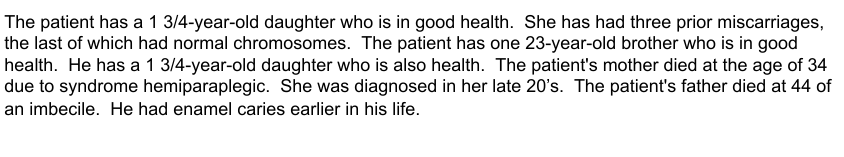}
\end{subfigure}%
\vspace{3pt}
\begin{subfigure}{.5\textwidth}
  \centering
  \includegraphics[width=.8\linewidth]{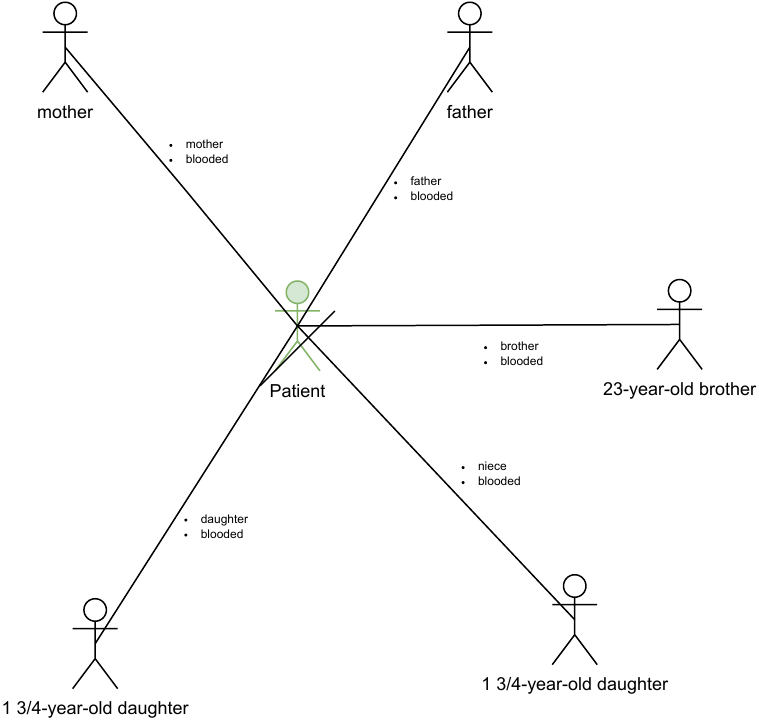}
  \caption{}
  \label{figure_family_member}
\end{subfigure}%
\begin{subfigure}{.5\textwidth}
  \centering
  \includegraphics[width=.8\linewidth]{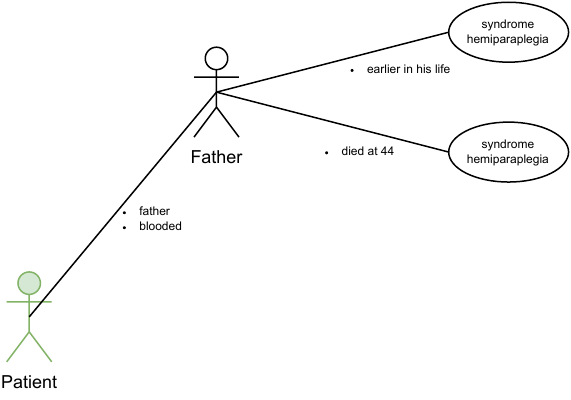}
  \caption{}
  \label{figure_family_graph}
\end{subfigure}
\caption{An example of \emph{document-to-graph} for family history extraction. A family history graph is build based on text shown on the upper part. For the sake of brevity, only clinical conditions relating to the father is shown in~\ref{figure_family_graph}.}
\label{figure-unified-framework}
\end{figure}

We propose to frame family history extraction as a \emph{document-to-graph} task.
That is, a family history graph---a mini version of knowledge graph about the patient's family history---should be inferred from text describing the patient's family history.
Constructing such a graph can be divided into two steps.
First, nodes that represent family members need to be constructed. 
Once a person is mentioned in text, a new node will be added to the graph.
By default, there is always a node representing the patient itself.
If the relation of identified family member and the patient is clearly described in text, we connect two nodes using the edge labeled with normalized relation (e.g., father, uncle, niece).
In some cases, the relation between family member and the patient needs to be inferred via the third family member.
Taking the sentence `The patient has ... brother ... he has a 1 3/4-year-old daughter ...' in Figure~\ref{figure-unified-framework} as an example, there are two family members mentioned: `brother' and `daughter'.
The relationship between these two family members is straightforward due to the existence of the linguistic pattern (i.e., X has Y).
We need to first connect these two nodes in the family graph and then infer the relationship between the patient and the family member with the help of coreference resolution (i.e., `he' refers to the brother).
Other attributes, such as side of family (maternal and paternal) and whether the family member is blooded relative of the patient (blooded, half blooded, not blooded) can be represented using different types of edges in this graph if these information is explicitly described in text (See example in Figure~\ref{figure_family_member}). 
Once we build the graph connecting all family members with the patient, the next step is recognizing all clinical conditions and linking them to the corresponding family member.
Attributes such as 'age of onset of diagnosis', 'age of death' and `uncertainty' are added to the edges if they are mentioned explicitly (See example in Figure~\ref{figure_family_graph}).


We believe the framework enables the usage of other resources to conduct transfer learning for family history extraction.
For example, the model may learn to recognize person mentions (and their attributes) and to identify the relationship between different persons via training on other types of narratives talking about interactions between multiple person. 
The setup in the second step is also very related to extracting disease names~\cite{dogan-nih-2014-jbi-ncbi-disease}; identifying social determinants of health~\cite{lybarger-washington-2021-jbi-sdoh}; and, recognizing adverse drug events~\cite{karimi-csiro-2015-jbi-cadec}, to name a few. 
The other benefit of this \emph{document-to-graph} formulation is that, once the graph is constructed, graph neural networks and knowledge graph embedding techniques may help with building family-aware patient representation, which enables downstream applications~\cite{chen-carter-2015-family-history,suissa-biu-2023-semantic-qa}.








\subsection{How can we integrate extracted family history into clinician's workflow?}

Once we obtain the data about the family history of the patient and a mini version of family history knowledge graph is constructed, the next question is how the extracted information should be presented to the clinician to provide decision support~\cite{taber-utah-2020-jamia-family-history,zajac-ucph-2023-tochi-clinician-ai}.
Lewis et al.~\cite{lewis-sfsu-2011-bicob-family-history} experiment with different visualization schemes---disease-centered view or family member centered view.
However, it is unclear how the clinician interacts with different views and how the interaction might affect clinician's perception toward these outputs.
To help clinicians to take advantage of extracted family history information and achieve user satisfaction, we need to understand the extend of clinicians information needs~\cite{alafaireet-belden-2017-medical-search}.
We believe extracted family history information may support decision making of the clinician in the following two scenarios:
\begin{itemize}
    \item Extracted family history information may help the clinician identify the health risk of the patient during the interaction with the patient.
    
    When the clinician evaluates the health condition of the patient, there are usually several considerations: a) reason for examination; b) examination or laboratory test results; c) patient medical history; and d) other factors such as family history, environmental risks and behavioral factor.
    On one hand, the clinician may be biased towards these recent obvious factors such as a) and b); on the other hand, the family history (and patient medical history) may be collected via multiple interactions between the patient and different health providers.
    The clinician may overlook hidden connections between these factors as these history information may be buried in various places, and automated extracted information may help identifying the health risk. 
    For example, if the patient has family history of `skin cancer' and an abnormal liver function test result, the clinician may consider skin checks and examine any lesions of concern if the check has not been done for more than 12 months.
    
    \item Extracted family history information may help with finding patients with specific clinical conditions, especially with particular familial disease history~\cite{dai-csiro-2021-cikm-searchehr,garcelon-upairs-2017-jamia-family-history}.
    
    In clinical trials which test a novel drug on people for safety and effectiveness, recruiting enough eligible patients is often a challenging job for the clinician. 
    The recruiting job can be more challenging if the trial aims to test novel preventative health screening or lifestyle.
    For example, the clinician may want to find healthy people who do not have diabetes but have a familial disease history of diabetes for testing low-carb diet.
    Conventional keyword querying over clinical notes is not effective as it does not distinguish different types of medical history.
    Information retrieval system, enhanced with extracted family history, may help with finding suitable participants.
    Another example of making use of extracted family history for finding patients is identifying patients who met evidence-based criteria for genetic testing~\cite{shi-utah-2022-jmir-family-history}.
    Fiol et al.~\cite{del-fiol-utah-2020-clinical-decision-support} utilize structured cancer family history data available in EHRs to identify unaffected patients who meet current guidelines for cancer genetic testing.
    We believe family history extracted from free-text clinical notes can enhance structured data and help clinician better refer patients to genetic services.
    
    
    %
    
    
    
    
\end{itemize}


\section{Summary}
We survey the literature on (1) what techniques from the NLP field are  developed to utilise digital health records for identification of risks of familial diseases, and (2) what resources are available to enable the development of family history extraction models. 

We identify that rule-based methods are heavily investigated and are still actively used for family history extraction, even though more recent efforts are focused on developing deep neural models that are based on large-scale pre-trained language models. Additionally, we identify the areas where more research is needed to unlock the value of patients' records, including collecting data about family history; transfer learning from other NLP resources under the {\em document-to-graph} task formulation; and, integrating extracted family history into clinician's workflow.



\paragraph*{Acknowledgements}
This work is supported by the CSIRO's Precision Health Future Science Platform (FSP).

\bibliographystyle{elsarticle-num} 
\bibliography{main}

\begin{thebibliography}{10}
\expandafter\ifx\csname url\endcsname\relax
  \def\url#1{\texttt{#1}}\fi
\expandafter\ifx\csname urlprefix\endcsname\relax\def\urlprefix{URL }\fi
\expandafter\ifx\csname href\endcsname\relax
  \def\href#1#2{#2} \def\path#1{#1}\fi

\bibitem{harris-2018-aged-care}
A.~Harris, A.~Sharma, {Estimating the future health and aged care expenditure
  in Australia with changes in morbidity}, PloS one 13 (2018).

\bibitem{brenner-stock-2014-colorectal-cancer}
H.~Brenner, C.~Stock, M.~Hoffmeister, {Effect of screening sigmoidoscopy and
  screening colonoscopy on colorectal cancer incidence and mortality:
  systematic review and meta-analysis of randomised controlled trials and
  observational studies}, The BMJ 348 (2014).

\bibitem{kaphingst-utah-2021-bmc-genetics}
K.~A. Kaphingst, W.~K. Kohlmann, R.~Chambers, M.~S. Goodman, R.~L. Bradshaw,
  P.~A. Chan, D.~Chavez-Yenter, S.~V. Colonna, W.~F. Espinel, J.~N. Everett,
  A.~Gammon, E.~Goldberg, J.~González, K.~J. Hagerty, R.~Hess, K.~Kehoe,
  C.~Kessler, K.~E. Kimball, S.~Loomis, T.~R. Martinez, R.~Monahan, J.~D.
  Schiffman, D.~Temares, K.~Tobik, D.~W. Wetter, D.~M. Mann, K.~Kawamoto, G.~D.
  Fiol, S.~S. Buys, O.~Ginsburg, {Comparing models of delivery for cancer
  genetics services among patients receiving primary care who meet criteria for
  genetic evaluation in two healthcare systems: BRIDGE randomized controlled
  trial}, BMC Health Services Research 21 (2021).

\bibitem{barrett-connor-1984-family-history}
E.~Barrett-Connor, K.~Khaw, {Family history of heart attack as an independent
  predictor of death due to cardiovascular disease}, Circulation (1984).

\bibitem{murff-byrne-2004-family-history}
D.~B. H.~Murff, S.~Syngal, {Cancer risk assessment: quality and impact of the
  family history interview}, American journal of preventive medicine (2004).

\bibitem{bill-minnesota-2014-amia-family-history}
R.~Bill, S.~Pakhomov, E.~S. Chen, T.~J. Winden, E.~W. Carter, G.~B. Melton,
  {Automated extraction of family history information from clinical notes}, in:
  AMIA Annual Symposium Proceedings, 2014 of Conference.

\bibitem{zhou-lu-2014-family-relatives}
L.~Zhou, Y.~Lu, C.~Vitale, P.~Mar, F.~Chang, N.~Dhopeshwarkar, R.~Rocha,
  {Representation of information about family relatives as structured data in
  electronic health records}, Applied clinical informatics (2014).

\bibitem{shen-mayo-2021-jmir-n2c2-2019-fh}
F.~Shen, S.~Liu, S.~Fu, Y.~Wang, S.~Henry, O.~Uzuner, H.~Liu, {Family History
  Extraction From Synthetic Clinical Narratives Using Natural Language
  Processing: Overview and Evaluation of a Challenge Data Set and Solutions for
  the 2019 National NLP Clinical Challenges (n2c2)/Open Health Natural Language
  Processing (OHNLP) Competition}, JMIR Medical Informatics 9 (2021).

\bibitem{rama-oslo-2018-louhi-family-history}
{\O}.~N. Taraka~Rama, P{\aa}l~Brekke, L.~{\O}vrelid, {Iterative development of
  family history annotation guidelines using a synthetic corpus of clinical
  text}, in: Proceedings of the Ninth International Workshop on Health Text
  Mining and Information Analysis, 2018 of Conference.

\bibitem{fuchs-giovannucci-1994-family-history}
C.~Fuchs, E.~Giovannucci, G.~Colditz, D.~Hunter, F.~Speizer, W.~Willett, {A
  prospective study of family history and the risk of colorectal cancer}, The
  New England journal of medicine (1994).

\bibitem{scheuner-wang-1997-family-history}
M.~Scheuner, S.~Wang, L.~Raffel, S.~K. Larabell, J.~Rotter, {Family history: a
  comprehensive genetic risk assessment method for the chronic conditions of
  adulthood}, American journal of medical genetics (1997).

\bibitem{guttmacher-collins-2004-family-history}
A.~Guttmacher, F.~Collins, R.~Carmona, {The family history--more important than
  ever}, The New England journal of medicine (2004).

\bibitem{del-fiol-utah-2020-clinical-decision-support}
G.~D. Fiol, W.~K. Kohlmann, R.~L. Bradshaw, C.~R. Weir, M.~C. Flynn, R.~Hess,
  J.~D. Schiffman, C.~J. Nanjo, K.~Kawamoto, {Standards-Based Clinical Decision
  Support Platform to Manage Patients Who Meet Guideline-Based Criteria for
  Genetic Evaluation of Familial Cancer}, JCO Clinical Cancer Informatics 4
  (2020).

\bibitem{polubriaginof-columbia-2015-amia-family-history}
N.~P.~T. Fernanda~Polubriaginof, D.~K. Vawdrey, {An assessment of family
  history information captured in an electronic health record}, in: AMIA Annual
  Symposium Proceedings, 2015 of Conference.

\bibitem{wang-mayo-2017-amia-family-history}
Y.~Wang, L.~Wang, M.~Rastegar-Mojarad, S.~Liu, F.~Shen, H.~Liu, {Systematic
  Analysis of Free-Text Family History in Electronic Health Record}, AMIA
  Summits on Translational Science Proceedings 2017 (2017).

\bibitem{permuth-egan-2009-familial-cancer}
J.~Permuth-Wey, K.~M. Egan, {Family history is a significant risk factor for
  pancreatic cancer: results from a systematic review and meta-analysis},
  Familial cancer 8 (2009).

\bibitem{patel-gonen-2012-genetic-profiling}
J.~P. Patel, M.~Gonen, M.~E. Figueroa, H.~Fernandez, Z.~Sun, J.~Racevskis,
  P.~V. Vlierberghe, I.~Dolgalev, S.~Thomas, O.~Aminova, {Prognostic relevance
  of integrated genetic profiling in acute myeloid leukemia}, New England
  Journal of Medicine 366 (2012).

\bibitem{roberts-fushman-2020-trec-precision-medicine}
K.~Roberts, D.~Demner-Fushman, E.~M. Voorhees, S.~Bedrick, W.~R. Hersh,
  {Overview of the TREC 2020 Precision Medicine Track}, in: TREC, 2020 of
  Conference.

\bibitem{milne-auckland-2009-family-hisory}
B.~J. Milne, A.~Caspi, H.~Harrington, R.~Poulton, M.~Rutter, T.~E. Moffitt,
  {Predictive value of family history on severity of illness: the case for
  depression, anxiety, alcohol dependence, and drug dependence}, Archives of
  general psychiatry 66 (2009).

\bibitem{claassen-vumc-2010-bmc-family-history}
L.~Claassen, L.~Henneman, A.~C. J.~W. Janssens, M.~Wijdenes-Pijl, N.~Qureshi,
  F.~M. Walter, P.~W. Yoon, D.~R.~M. Timmermans, {Using family history
  information to promote healthy lifestyles and prevent diseases; a discussion
  of the evidence}, BMC Public Health 10 (2010).

\bibitem{mehrabi-mayo-2015-family-history}
S.~Mehrabi, A.~Krishnan, A.~M. Roch, H.~Schmidt, D.~Li, J.~Kesterson,
  C.~Beesley, P.~Dexter, M.~Schmidt, M.~Palakal, {Identification of patients
  with family history of pancreatic cancer-Investigation of an NLP System
  Portability}, Studies in health technology and informatics 216 (2015).

\bibitem{taber-utah-2020-jamia-family-history}
P.~Taber, P.~Ghani, J.~D. Schiffman, W.~K. Kohlmann, R.~Hess, V.~Chidambaram,
  K.~Kawamoto, R.~G. Waller, D.~A. Borbolla, G.~D. Fiol, C.~R. Weir,
  {Physicians' strategies for using family history data: having the data is not
  the same as using the data}, JAMIA open 3 (2020).

\bibitem{chen-vermont-2012-amia-family-history}
E.~S. Chen, G.~B. Melton, T.~E. Burdick, P.~T. Rosenau, I.~N. Sarkar,
  {Characterizing the use and contents of free-text family history comments in
  the Electronic Health Record}, in: AMIA Annual Symposium Proceedings, 2012 of
  Conference.

\bibitem{lewis-ibm-2011-ieee-parsing-family-history}
D.~F.~G. Neal~Lewis, H.~Yang, {Dependency Parsing for Extracting Family
  History}, 2011 IEEE First International Conference on Healthcare Informatics,
  Imaging and Systems Biology (2011).

\bibitem{balabaeva-kovalchuk-2020-family-disease-tree}
K.~Balabaeva, S.~Kovalchuk, {Experiencer Detection and Automated Extraction of
  a Family Disease Tree from Medical Texts in Russian Language}, in:
  International Conference on Computational Science, 2020 of Conference.

\bibitem{lewis-sfsu-2011-bicob-family-history}
D.~G. Neal~Lewis, H.~Yang, {Extracting Family History Diagnosis from Clinical
  Texts}, in: Proceedings of the {ISCA} 3rd International Conference on
  Bioinformatics and Computational Biology, 2011 of Conference.

\bibitem{dai-csiro-2021-cikm-searchehr}
X.~Dai, M.~Rybinski, S.~Karimi, {SearchEHR: A Family History Search System for
  Clinical Decision Support}, in: Proceedings of the 30th ACM International
  Conference on Information \& Knowledge Management, 2021 of Conference.

\bibitem{yang-florida-2020-jmir-family-history}
X.~Yang, H.~Zhang, X.~He, J.~Bian, Y.~Wu, {Extracting Family History of
  Patients From Clinical Narratives: Exploring an End-to-End Solution With Deep
  Learning Models}, JMIR Medical Informatics 8 (2020).

\bibitem{mowery-kawamoto-2019-amia-family-history}
D.~L. Mowery, K.~Kawamoto, R.~Bradshaw, W.~K. Kohlmann, J.~D. Schiffman, C.~R.
  Weir, D.~A. Borbolla, W.~W. Chapman, G.~D. Fiol, {Determining Onset for
  Familial Breast and Colorectal Cancer from Family History Comments in the
  Electronic Health Record}, AMIA Joint Summits on Translational Science
  proceedings 2019 (2019).

\bibitem{wan-ge-2022-family-history-mood-disorder}
C.~Wan, X.~Ge, J.~Wang, X.~Zhang, Y.~Yu, J.~Hu, Y.~Liu, H.~Ma, {Identification
  and Impact Analysis of Family History of Psychiatric Disorder in Mood
  Disorder Patients With Pretrained Language Model}, Frontiers in Psychiatry 13
  (2022).

\bibitem{shi-utah-2022-jmir-family-history}
J.~Shi, K.~L. Morgan, R.~L. Bradshaw, S.-H. Jung, W.~Kohlmann, K.~A. Kaphingst,
  K.~Kawamoto, G.~D. Fiol, {Identifying Patients Who Meet Criteria for Genetic
  Testing of Hereditary Cancers Based on Structured and Unstructured Family
  Health History Data in the Electronic Health Record: Natural Language
  Processing Approach}, JMIR Medical Informatics 10 (2022).

\bibitem{friedlin-mcdonald-2006-amia-family-history}
J.~Friedlin, C.~J. McDonald, {Using a natural language processing system to
  extract and code family history data from admission reports}, in: AMIA Annual
  Symposium Proceedings, 2006 of Conference.

\bibitem{goryachev-harvard-2008-amia-family-history}
H.~K. Sergey~Goryachev, Q.~Zeng-Treitler, {Identification and Extraction of
  Family History Information from Clinical Reports}, in: AMIA Annual Symposium
  Proceedings, 2008 of Conference.

\bibitem{rastegar-mayo-2018-bcb-biocreative-2018}
M.~Rastegar-Mojarad, S.~Liu, Y.~Wang, N.~Afzal, L.~Wang, F.~Shen, S.~Fu,
  H.~Liu, {BioCreative/OHNLP Challenge 2018}, in: Proceedings of the 2018 ACM
  International Conference on Bioinformatics, Computational Biology, and Health
  Informatics, 2018 of Conference.

\bibitem{azab-umich-2019-emnlp-family-history}
M.~Azab, S.~Dadian, V.~Nastase, L.~An, R.~Mihalcea, {Towards extracting medical
  family history from natural language interactions: A new dataset and
  baselines}, in: Proceedings of the 2019 Conference on Empirical Methods in
  Natural Language Processing and the 9th International Joint Conference on
  Natural Language Processing (EMNLP-IJCNLP), 2019 of Conference.

\bibitem{bodenreider-2004-umls}
O.~Bodenreider, {The Unified Medical Language System (UMLS): integrating
  biomedical terminology}, Nucleic acids research 32 (2004).

\bibitem{almeida-matos-2020-sac-family-history}
J.~{a}o Rafael~Almeida, S.~Matos, {Rule-based extraction of family history
  information from clinical notes}, in: Proceedings of the 35th Annual ACM
  Symposium on Applied Computing, 2020 of Conference.

\bibitem{manning-stanford-2014-acl-corenlp}
C.~D. Manning, M.~Surdeanu, J.~Bauer, J.~R. Finkel, S.~Bethard, D.~McClosky,
  \href{https://aclanthology.org/P14-5010.pdf}{{The Stanford CoreNLP Natural
  Language Processing Toolkit}}, in: Proceedings of 52nd annual meeting of the
  association for computational linguistics: system demonstrations, 2014 of
  Conference.
\newline\urlprefix\url{https://aclanthology.org/P14-5010.pdf}

\bibitem{zhang-stanford-2021-jamia-stanza}
Y.~Zhang, Y.~Zhang, P.~Qi, C.~D. Manning, C.~P. Langlotz, {Biomedical and
  clinical English model packages for the Stanza Python NLP library}, Journal
  of the American Medical Informatics Association 28 (2021).

\bibitem{dai-2019-bmc-family-member}
H.-J. Dai, {Family member information extraction via neural sequence labeling
  models with different tag schemes}, BMC medical informatics and decision
  making 19 (2019).

\bibitem{dai-nkust-2020-jmir-family-history}
C.~N. Hong-Jie~Dai, You-Qian~Lee, J.~Jonnagaddala, {Family History Information
  Extraction With Neural Attention and an Enhanced Relation-Side Scheme:
  Algorithm Development and Validation}, JMIR medical informatics 8 (2020).

\bibitem{rybinski-csiro-2021-jmir-family-history}
M.~Rybinski, X.~Dai, S.~Singh, S.~Karimi, A.~Nguyen, {Extracting Family History
  Information From Electronic Health Records: Natural Language Processing
  Analysis}, JMIR Medical Informatics 9 (2021).

\bibitem{kim-musc-2021-jmir-family-history}
I.~R.~L. Youngjun~Kim, Paul M.~Heider, S.~M. Meystre, {A Hybrid Model for
  Family History Information Identification and Relation Extraction:
  Development and Evaluation of an End-to-End Information Extraction System},
  JMIR Medical Informatics 9 (2021).

\bibitem{shi-hit-2019-bmc-family-history}
X.~Shi, D.~Jiang, Y.~Huang, X.~Wang, Q.~Chen, J.~Yan, B.~Tang, {Family history
  information extraction via deep joint learning}, BMC medical informatics and
  decision making 19 (2019).

\bibitem{gururangan-allenai-2020-acl-dapt}
S.~Gururangan, A.~Marasovi{\'c}, S.~Swayamdipta, K.~Lo, I.~Beltagy, D.~Downey,
  N.~A. Smith, {Don't Stop Pretraining: Adapt Language Models to Domains and
  Tasks}, in: Proceedings of the 58th Annual Meeting of the Association for
  Computational Linguistics, 2020 of Conference.

\bibitem{pruksachatkun-nyu-2020-acl-intermediate-task}
Y.~Pruksachatkun, J.~Phang, H.~Liu, P.~M. Htut, X.~Zhang, R.~Y. Pang, C.~Vania,
  K.~Kann, S.~R. Bowman, {Intermediate-Task Transfer Learning with Pretrained
  Models for Natural Language Understanding: When and Why Does It Work?}, in:
  Proceedings of the 58th Annual Meeting of the Association for Computational
  Linguistics, 2020 of Conference.

\bibitem{devlin-google-2019-naacl-bert}
K.~L. Jacob~Devlin, Ming-Wei~Chang, K.~Toutanova, {BERT: Pre-training of deep
  bidirectional transformers for language understanding}, in: Proceedings of
  the 2019 Conference of the North {A}merican Chapter of the Association for
  Computational Linguistics: Human Language Technologies, Volume 1 (Long and
  Short Papers), 2019 of Conference.

\bibitem{silva-rafael-2020-jmir-family-history}
J.~F. Silva, J.~R. Almeida, S.~Matos, {Extraction of family history information
  from clinical notes: deep learning and heuristics approach}, JMIR medical
  informatics 8 (2020).

\bibitem{komninos-manandhar-2016-naacl-dependency-embeddings}
A.~Komninos, S.~Manandhar, {Dependency based embeddings for sentence
  classification tasks}, in: Proceedings of the 2016 conference of the North
  American chapter of the association for computational linguistics: human
  language technologies, 2016 of Conference.

\bibitem{peters-allenai-2018-naacl-elmo}
M.~E. Peters, M.~Neumann, M.~Iyyer, M.~Gardner, C.~Clark, K.~Lee,
  L.~Zettlemoyer, {Deep contextualized word representations}, in: Proceedings
  of the 2018 Conference of the North {A}merican Chapter of the Association for
  Computational Linguistics: Human Language Technologies, Volume 1 (Long
  Papers), 2018 of Conference.

\bibitem{zhan-hit-2021-jmir-family-history}
K.~Zhan, W.~Peng, Y.~Xiong, H.~Fu, Q.~Chen, X.~Wang, B.~Tang, {Novel
  Graph-Based Model With Biaffine Attention for Family History Extraction From
  Clinical Text: Modeling Study}, JMIR medical informatics 9 (2021).

\bibitem{ma-hovy-2016-acl-ner}
X.~Ma, E.~Hovy, {End-to-end Sequence Labeling via Bi-directional
  LSTM-CNNs-CRF}, in: Proceedings of the 54th Annual Meeting of the Association
  for Computational Linguistics (Volume 1: Long Papers), 2016 of Conference.

\bibitem{dozat-manning-2017-iclr-biaffine}
T.~Dozat, C.~D. Manning, {Deep biaffine attention for neural dependency
  parsing}, in: International Conference on Learning Representations, 2017 of
  Conference.

\bibitem{gao-wisc-2021-jamia-review-clinical-nlp}
Y.~Gao, D.~Dligach, L.~Christensen, S.~Tesch, R.~Laffin, D.~Xu, T.~Miller,
  O.~Uzuner, M.~M. Churpek, M.~Afshar, {A Scoping Review of Publicly Available
  Language Tasks in Clinical Natural Language Processing}, Journal of the
  American Medical Informatics Association 29 (2021).

\bibitem{spasic-nenadic-2020-jmir-clinical-review}
I.~Spasic, G.~Nenadic, {Clinical text data in machine learning: systematic
  review}, JMIR medical informatics 8 (2020).

\bibitem{clift-mayo-2022-family-history}
K.~Clift, S.~Macklin-Mantia, M.~Barnhorst, L.~Millares, Z.~King, A.~Agarwal,
  R.~J. Presutti, {Comparison of a Focused Family Cancer History Questionnaire
  to Family History Documentation in the Electronic Medical Record}, Journal of
  Primary Care \& Community Health 13 (2022).

\bibitem{brekke-rama-2021-synthetic-data}
P.~Brekke, T.~Rama, I.~Pil{\'a}n, {\O}.~Nytr{\o}, L.~{\O}vrelid, {Synthetic
  data for annotation and extraction of family history information from
  clinical text}, Journal of Biomedical Semantics (2021).

\bibitem{laranjo-mq-2018-jamia-chatbot}
L.~Laranjo, A.~G. Dunn, H.~L. Tong, A.~B. Kocaballi, J.~Chen, R.~Bashir,
  D.~Surian, B.~Gallego, F.~Magrabi, A.~Y. Lau, E.~Coiera, {Conversational
  agents in healthcare: a systematic review}, Journal of the American Medical
  Informatics Association 25 (2018).

\bibitem{dogan-nih-2014-jbi-ncbi-disease}
R.~L. Rezarta Islamaj~Do{\u{g}}an, Z.~Lu, {NCBI disease corpus: A resource for
  disease name recognition and concept normalization}, Journal of biomedical
  informatics 47 (2014).

\bibitem{lybarger-washington-2021-jbi-sdoh}
M.~O. Kevin~Lybarger, M.~Yetisgen, {Annotating social determinants of health
  using active learning, and characterizing determinants using neural event
  extraction}, Journal of Biomedical Informatics 113 (2021).

\bibitem{karimi-csiro-2015-jbi-cadec}
M.~K. Sarvnaz~Karimi, Alejandro Metke-Jimenez, C.~Wang, {CADEC: A corpus of
  adverse drug event annotations}, Journal of biomedical informatics 55 (2015).

\bibitem{chen-carter-2015-family-history}
E.~S. Chen, E.~W. Carter, T.~J. Winden, I.~N. Sarkar, Y.~Wang, G.~B. Melton,
  {Multi-source development of an integrated model for family health history},
  Journal of the American Medical Informatics Association (2015).

\bibitem{suissa-biu-2023-semantic-qa}
M.~Z.-G. Omri~Suissa, A.~Elmalech, {Question answering with deep neural
  networks for semi-structured heterogeneous genealogical knowledge graphs},
  Semantic Web 14 (2023).

\bibitem{zajac-ucph-2023-tochi-clinician-ai}
H.~D. Zajac, D.~Li, X.~Dai, J.~F. Carlsen, F.~Kensing, T.~O. Andersen,
  \href{https://dl.acm.org/doi/abs/10.1145/3582430}{{Clinician-facing AI in the
  Wild: Taking Stock of the Sociotechnical Challenges and Opportunities for
  HCI}}, ACM Transactions on Computer-Human Interaction (2023).
\newline\urlprefix\url{https://dl.acm.org/doi/abs/10.1145/3582430}

\bibitem{alafaireet-belden-2017-medical-search}
P.~Alafaireet, J.~Belden, M.~Botkin, K.~Kochendorfer, R.~Kruse, D.~Strecker,
  J.~Williams, {Embedding a Medical Search Engine Within an Electronic Health
  Record}, Missouri Medicine 114 (2017).

\bibitem{garcelon-upairs-2017-jamia-family-history}
N.~Garcelon, A.~Neuraz, V.~Benoit, R.~Salomon, A.~Burgun, {Improving a
  full-text search engine: the importance of negation detection and family
  history context to identify cases in a biomedical data warehouse}, Journal of
  the American Medical Informatics Association 24 (2017).

\end{thebibliography}

\end{document}